\documentclass[mlmain]{jmlr}

\usepackage{url}
\usepackage{xurl}
\usepackage{hyperref}
\usepackage{longtable}
\usepackage{booktabs}
\usepackage[load-configurations=version-1]{siunitx}

\jmlrvolume{}
\firstpageno{1}
\jmlrproceedings{PMLR}{Proceedings of Machine Learning Research}
\jmlryear{2026}
\jmlrworkshop{Symmetry and Geometry in Neural Representations}

\title[Neighbor Crowding Predicts Feature Stability in Sparse Autoencoders]{Neighbor Crowding Predicts Feature Stability in Sparse Autoencoders}

\author{\Name{Denis Federico Lim} \Email{denislim@alumni.usc.edu}\\
  \addr Korea Advanced Institute of Science and Technology (KAIST), Daejeon, South Korea
  \AND
  \Name{Naveen Dasari} \Email{nsd2139@engineering.upenn.edu}\\
  \addr University of Pennsylvania, Philadelphia, USA
  \AND
  \Name{George Flint} \Email{georgeflint@berkeley.edu}\\
  \addr University of Cambridge, Cambridge, UK
  \AND
  \Name{Kevin Zhu} \Email{zhu502846@berkeley.edu}\\
  \addr University of California, Berkeley, USA
  }

\begin{document}

\maketitle

\begin{abstract}
Sparse Autoencoders (SAEs) offer an approach to identify language model representations by breaking apart polysemantic features into more sparse, interpretable ones. Yet, it is challenging to identify which of the learned features are actually stable, rather than artifacts of initialization. One approach is to train multiple SAE seeds, and compare the learned features across seeds by cosine similarity, but this is notably costly. This raises the question of whether stable features can be identified without the computational expense of training multiple seeds. In this work, we examine cross-seed feature stability across differing levels of training tokens, and demonstrate that in TopK SAEs, it is highly predictable using six computationally cheap statistics derived from a single training run. Neighbor crowding, a geometric property of the dictionary, is the strongest single predictor, with features that sit in dense regions of the unit sphere more likely to be stable than isolated ones.

\paragraph{Code:} \url{https://github.com/DenisLim95/sparse-autoencoder-stability-project}

\begin{keywords}
sparse autoencoders, mechanistic interpretability, feature stability, TopK, dictionary learning
\end{keywords}

\end{abstract}

\section{Introduction}
\label{sec:intro}
In modern deep learning, sparse autoencoders (SAEs) have developed into a leading approach for decomposing dense, polysemantic representations into a higher-dimensional space of sparse, more interpretable features \citep{bereska2024mechanistic,cunningham2023sparse,sharkey2022interim,bricken2023towards, superposition}. Neural architectures often represent inputs in low-dimensional subspaces whose appropriate dimensionality is task-dependent \citep{badre2020dimensionality}. Any such subspace of dimension $\geq 2$, by definition, cannot be spanned by a single unique orthonormal basis \citep{waldron_linalg}. Previous research suggests a neurobiological analogy: a single population-level geometry can be determined by different sets of individual underlying neurons \citep{kriegeskorte2021neural}. A single decoder column within an SAE, known as a feature, is analogous in theory to a neuron, and exhibits a similar property in practice: across two seeds, an SAE trained on Llama 3 8B showed only $30\%$ shared features, suggesting baseline instability \citep{paulo2025sparse}.

Feature stability is particularly important, in that, if it shows stability across different initial seeds, properties of that feature broadly generalize beyond a single training run. This allows an observer to credibly develop understandings about the internal representation of large language models (LLMs). This raises the following question: how does feature stability change across TopK architecture and across increasing training tokens, and can stable features be predicted from properties available in a single training run without the need for retraining across seeds?
We make the following contributions:
\begin{itemize}
    \item \textbf{Cross-seed and budget stability.} We train three SAE seeds on Pythia-70m, at TopK sparsity levels $k \in \{64, 128, 256\}$. We categorize how feature stability varies across seeds and different training token budgets.
    \item \textbf{Single-run predictor.} We show that six cheap statistics from one single run are enough to train a logistic regression model to predict feature stability, which avoids need for expensive multi-seed runs.
    \item \textbf{Neighbor Crowding is the strongest single stability predictor.} We show that neighbor crowding predicts feature stability better on its own than any of the five other statistics.
    \item \textbf{Cross-scale and cross-family transfer.} We show that a classifier trained on Pythia-70m predicts stability on larger Pythia models nearly as well as classifiers trained on their own labels, and transfers to GPT-2 small, a different model family, once its statistics are standardized on the target model.
\end{itemize}

\section{Related Work}
\citet{paulo2025sparse} report that SAEs differing only in initialization share as few as $30\%$ of their features, and that TopK SAEs are more seed-dependent than $\ell_1$-regularized ReLU SAEs at matched sparsity. Directly relevant to the sweep reported here, they also find that cross-seed overlap decreases as $k$ increases and increases as training runs longer. \citet{gerasimov2026unstable} convert these population-level observations into a per-feature quantity by estimating each feature's reappearance probability across a large set of independently trained SAEs. In so doing, they report the same monotone relationship within TopK SAEs: increasing $k$ improves explained variance but raises the unstable fraction and lowers the stable fraction. Their labels require training many SAEs per model. They also show that a logistic regression on a feature's $d$-dimensional decoder vector, fitted on one seed, predicts instability on another. Such a classifier is tied to the width $d$ of the model it was fitted on and cannot be applied to a model of different width. We instead predict stability from six width-agnostic, interpretable statistics, which lets a classifier fitted on one model transfer to larger models and to another model family.

\citet{song2026mechanistic} propose that activation frequency is highly correlated with feature similarity across runs, with frequently activated features being more consistent. We take this as the baseline that our methods must beat, and we report our results as values of $\Delta$ compared against activation frequency alone.

\citet{lan2024quantifying} find that SAE feature spaces recur across models, especially in the middle layers. Additionally, \citet{son2025semantic} report that representations are shared across different scaled LLMs. These results motivate our cross-scale analysis; while they establish that there are consistent structures across model sizes, we ask if a stability classifier trained on one model can make predictions on bigger models of the same family and on a model from a different family.

\section{Methodology}
\subsection{SAE training setup}
\label{sec:training}
We run SAEs on Pythia-70m-deduped to avoid inflating activation frequency as it is one of the statistics we use for our classifier. Specifically, we use a TopK activation function and extract activation vectors after layer 3 ($d = 512$, dictionary containing $n = 8192$ features, a $16\times$ expansion of the residual stream) \citep{gao2024scaling,biderman2023pythia,makhzani_k_SAEs}. For some input activation $x$, we compute the features $f$ and the reconstruction $\hat{x}$ as:

\begin{align*}
f &= \text{TopK}(W_{enc}(x - b_{dec}) + b_{enc}) \\
\hat{x} &= W_{dec} f + b_{dec}
\end{align*}

Under the TopK regime, distinct from $\ell_1$-regularized SAEs, no sparsity penalty is applied, so the training objective is the standard reconstruction loss term $\mathcal{L}_{recon} = \mathbb{E}_x\left[\|x - \hat{x}\|_2^2\right]$ \citep{gao2024scaling}. We constrain the decoder vector columns to unit $\ell_2$ norm throughout training so that decoder columns encode pure feature directions and activation magnitudes are directly comparable across features and runs.

We train the SAEs across a sweep of features $k \in \{64, 128, 256\}$, for three random initialized seeds per $k$ ($42, 256, 1024$) and with six training-token checkpoints ($50$M, $100$M, $250$M, $500$M, $800$M, $1$B). Concretely, all nine runs are trained on the Pile Dataset \citep{gao2020pile}, and are optimized with Adam at a learning rate of $10^{-3}$ on batches of $256 \times 128 = 32{,}768$ tokens per step. Differences in the resulting dictionary can be attributable to differences in initialization seeds alone, as training data, data order and hyperparameters are held constant across runs. A shared data order likely makes seeds easier to match, which we discuss in \sectionref{sec:limitations}. We additionally adopt the two standard mitigations against dictionary collapse under a TopK objective, both introduced by \citet{gao2024scaling}: encoder weights are initialized to be tied to the transpose of the decoder, and an auxiliary reconstruction term (AuxK, with coefficient $0.03125$ over the top $512$ inactive latents) periodically revives latents that have stopped firing. For the cross-scale evaluation, we train three SAEs at $k=64$ on Pythia-160m-deduped at layer 6, with the same three seeds and training tokens, and repeat this on Pythia-410m-deduped at layer 12 ($d = 1024$, $n = 16{,}384$) to 500M tokens. For the cross-family evaluation, we do the same on GPT-2 small at layer 6 ($d = 768$, $n = 12{,}288$), which matches Pythia-160m's width and depth, using activations on the same Pile text, to 500M tokens. Finally, to check sensitivity to the number of seeds, we train five Pythia-70m SAEs at $k=64$ (the original three seeds plus $137$ and $512$) from scratch on the same data stream, to 500M tokens.

\subsection{Stability labels}
We adopt the labeling convention of \citet{gerasimov2026unstable}, comparing features across seeds by their decoder weights. For two SAEs with decoder matrices $D_1, D_2 \in \mathbb{R}^{n \times d}$, we $\ell_{2}$ normalize the decoder vector of each feature and compute the pairwise cosine similarity matrix between the two dictionaries. Following \citet{gerasimov2026unstable}, we use many-to-one argmax matching and consider two features to be the same across seeds when the best match passes threshold $\theta$=0.7.

For an anchor seed, a feature is labeled \emph{stable} if its best match in every other seed (two in our main experiments, four in the five-seed check) has cosine similarity $\geq \theta$, and \emph{unstable} if it has no match above $\theta$ in any of them. Features matching in some but not all other seeds are left unlabeled and excluded from the classifier, which is fit only on labeled features that also clear the firing floor (\sectionref{sec:predictors}). Reported stable fractions are over all $n$ latents unless marked live-only.

\subsection{Single-run predictors}
\label{sec:predictors}
We evaluate six statistics from a single run over a held-out set of 1.31M tokens as predictors of feature stability:

\paragraph{Activation frequency.} The fraction of times a feature fires over the evaluation tokens, measured in $\log_{10}$ units. This is the baseline our results compare against.

\paragraph{Mean activation.} The average activation magnitude,
conditional on the feature firing, in $\log_{10}$ units.

\paragraph{Neighbor crowding.} The mean cosine similarity between a feature's decoder column and its ten nearest neighbors within the same SAE.

\paragraph{Reconstruction contribution.} The mean increase in reconstruction mean-squared error (MSE) per token when zeroing the feature after encoding, conditional on the feature firing.

\paragraph{Encoder column norm.} The magnitude of the response.

\paragraph{Encoder bias ($b_{\mathrm{enc}}$).} The activation threshold.

The conditional statistics are not well-calculated when features are completely or frequently inactive. To mitigate this, we set a floor threshold of 100 firings to help prevent the classifier from succeeding merely by identifying dead features. All reported figures use this threshold.

\paragraph{Standardization.}
We standardize by converting each statistic to a $z$-score. Given that the statistics live on very different orders of magnitude, standardization ensures the $\ell_2$ penalty applies evenly across predictors.

\paragraph{Classifier.}
We fit an $\ell_2$-regularized logistic regression with class weights inversely proportional to class frequency, and report mean AUROC over stratified five-fold cross-validation (\appendixref{apd:training}).

\subsection{Evaluation and generalization}
Our main metric is the stability prediction accuracy gain as compared to the baseline provided by activation frequency alone:
\[
\Delta = \text{AUROC}(\text{all six stats}) - \text{AUROC}(\text{activation frequency alone})
\]

We re-ran the cross-validation process 1000 times with permutations of the stability labels and verified classifier accuracy results in random chance (mean AUROC $0.500$, $0.499$ and $0.499$ at $k=64$, $128$ and $256$).

We test three levels of transfer conditions. First we do a cross-seed transfer test, where the classifier is trained on one SAE seed and evaluated on a different seed's dictionary. Second, we test cross-budget transfer, fitting the classifier on an early checkpoint (Pythia-70m) and evaluating on a later one. Thirdly we test cross-scale transfer, fitting the predictor on Pythia-70m and testing against dictionaries trained on Pythia-160m and Pythia-410m, and cross-family transfer to dictionaries trained on GPT-2 small.

For the cross-scale condition we take activations at a matched relative depth, half of each model's layers, so that layer position is comparable across models of differing depth, and we hold the dictionary at a $16\times$ expansion of each model's residual stream with $k$ fixed at $64$.

For all transfer evaluations, we standardize the target's statistics with either the source's means and standard deviations (\emph{source scaler}) or the target's own (\emph{target scaler}), and compare both against a classifier cross-validated directly on the target's labels. AUROC depends only on how features are ranked, so the two scalers differ only in how much weight each statistic receives. Refitting the scaler needs the target's statistics, not its labels.

\section{Results}
\label{sec:results}
We find that, across sparsity levels $k \in \{64, 128, 256\}$, cross-seed feature stability differs in both absolute levels as well as the underlying shape across differing levels of training tokens. At the highest level of training tokens (1B), we report that $\approx 64\%$ of features are stable at $k=64$, $\approx 56\%$ at $k=128$, and near-zero at $k=256$ (\figureref{fig:stability_by_token_count}). The trajectories for feature stability for $k=64$ and $k=128$ reflect an S-curve shape across the training-token sweep, while the trajectory for $k=256$ is near-zero throughout the full range of tokens.

\begin{figure}[htbp]
\floatconts
  {fig:stability_by_token_count}
  {\caption{Both stability and reappearance probability generally increase relatively significantly for $k = 64, 128$ across training token sizes, but the curve is flatter for $k=256$ (a solid line indicates an all features curve; dashed, live-only).}}
  {\includegraphics[width=0.55\linewidth]{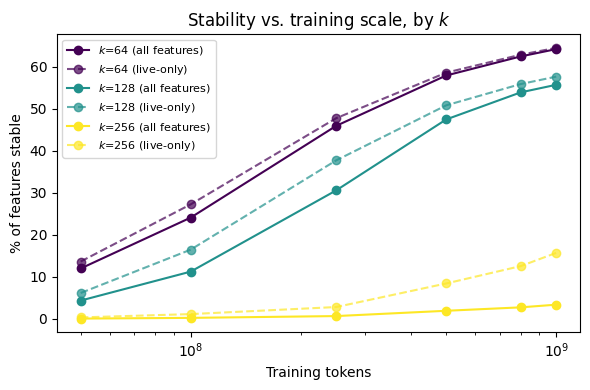}}
\end{figure}

\subsection{Stability is predictable from single-run statistics}
\label{sec:predictable}
We turn to whether cross-seed feature stability can be anticipated without training multiple seeds. Averaging over three different anchor seeds at $k=64$ and 1B training tokens, we find that the six single-run statistics predict feature stability with an AUROC of $0.869$. This provides a $\Delta = +0.259$, as compared against $0.610$ for activation frequency only. A classifier fitted on one seed's dictionary also transfers to another seed's, scoring within $0.0013$ of that seed's own ceiling. We additionally train a fresh five-seed run to 500M tokens and label it against four partners rather than two. Compared with labels from three of those seeds, $\Delta$ rises from $+0.265$ to $+0.291$, and the AUROC of neighbor crowding alone from $0.812$ to $0.857$, so predictability does not hinge on using three seeds.

\begin{table}[t]
  \centering
  \caption{Classifier accuracy at different $k$ values at 1B tokens, anchor seed 42. \emph{live}: percent of latents clearing the firing floor; $\Delta$: AUROC(all six) $-$ AUROC(frequency alone).}
  \label{tab:by_k}
  \begin{tabular}{lccccc}
    \toprule
    $k$ & live & stable & AUROC & frequency only & $\Delta$ \\
    \midrule
    64  & $97.3\%$ & $64.2\%$ & $0.868$ & $0.606$ & $+0.262$ \\
    128 & $89.4\%$ & $55.7\%$ & $0.863$ & $0.678$ & $+0.185$ \\
    256 & $21.4\%$ & $3.4\%$  & $0.708$ & $0.599$ & $+0.110$ \\
    \bottomrule
  \end{tabular}
\end{table}

\subsection{Neighbor crowding is the strongest single predictor}
On its own, neighbor crowding recovers the stability label at AUROC $0.816$, ahead of reconstruction contribution ($0.802$), mean activation ($0.780$) and activation frequency ($0.610$) (\tableref{tab:per_predictor} in \appendixref{apd:per_predictor}). Crowding and frequency together reach $0.818$, and all six give $0.869$. Crowding is not the only statistic the full classifier relies on. Dropping it costs $0.032$ AUROC, comparable to dropping mean activation ($0.037$), while dropping any other statistic costs less than $0.01$. Features sitting in dense regions of the unit sphere tend to recur across seeds, over isolated ones.

\subsection{Stability prediction transfers across model scale\texorpdfstring{ and family}{ and family}}
\label{sec:crossscale}
The classifier is able to predict stability for the target model and across scales. When fitted on Pythia-160m's own labels, the same six statistics predict feature stability with a score of 0.851, which also beats the activation frequency baseline of 0.573, resulting in an improvement of $\Delta$ = +0.278 (\tableref{tab:by_model} in \appendixref{apd:by_model}).

More interestingly, we evaluated cross-scale transferability by fitting a classifier on Pythia-70m and testing on Pythia-160m. We found it scores $0.846$ when scaler is refitted on the target, and $0.848$ when using source's original scaler, which means the penalty of transferring is roughly $0.006$, as compared to the ceiling of $0.852$.

The same holds at Pythia-410m, which has twice the source's width and four times its depth, where the 70m classifier scores $0.819$ against a $0.831$ ceiling (\tableref{tab:crossscale}). At a matched 500M-token budget, the transfer penalty grows only from $0.003$ at 160m to $0.012$ at 410m.

On GPT-2 small, a different model family, the 70m classifier scores $0.772$ against a $0.823$ ceiling when the scaler is refit on GPT-2's statistics, but only $0.414$ with the source scaler. A score below chance means the source scaler inverts the ranking rather than merely weakening it. Reconstruction contribution, a squared error, has $142\times$ the spread in GPT-2 as in Pythia-70m, so under 70m's standardization it dominates the score. Within GPT-2, neighbor crowding is again the strongest single predictor (AUROC $0.698$ against $0.637$ for mean activation, anchor seed 42).

\begin{table}[t]
  \centering
  \caption{Cross-scale and cross-family transfer from Pythia-70m to Pythia-160m, Pythia-410m and GPT-2 small at $k=64$, averaged (three source seeds by three target seeds). \emph{source scaler} reuses the source's per-statistic standardization; \emph{target scaler} refits it on the target's own statistics, which requires no labels; \emph{ceiling} is a classifier cross-validated on the target's own labels; \emph{frequency only} is the target's own baseline.}
  \label{tab:crossscale}
  \begin{tabular}{llccccc}
    \toprule
    target & tokens & source scaler & target scaler & ceiling & frequency only & labeled \\
    \midrule
    Pythia-160m & 500M & $0.841$ & $0.839$ & $0.842$ & $0.561$ & $9{,}419$ \\
    Pythia-160m & 1B   & $0.848$ & $0.846$ & $0.852$ & $0.571$ & $9{,}855$ \\
    Pythia-410m & 500M & $0.816$ & $0.819$ & $0.831$ & $0.556$ & $12{,}067$ \\
    GPT-2 small & 500M & $0.414$ & $0.772$ & $0.823$ & $0.494$ & $9{,}749$ \\
    \bottomrule
  \end{tabular}
\end{table}

\subsection{Confound analysis}
\label{sec:confound}
Relative to $k=64$ and $128$, stability at $k=256$ is near zero. Neither dead latents nor a miscalibrated $\theta$ explains this difference, since even among live latents only $\approx 16\%$ are stable at $k=256$, and near-duplicate features are rare at every $k$ (\appendixref{apd:k256}).

Neighbor crowding is a geometric property of a single dictionary, so it could in principle predict the label only because both reflect how densely features are packed on the unit sphere. To test this, we rotate the partner dictionary by a random orthogonal matrix, which preserves its geometry while destroying any correspondence between the two dictionaries. No rotated pairing clears $\theta = 0.7$, so to score this null we label as stable the same fraction of features as in the real labels, taking those with the highest best-match cosine to the rotated partner. Against these labels, neighbor crowding predicts at chance (AUROC $\approx 0.501$), so its predictive power reflects genuine cross-seed correspondence rather than sphere density. Its predictive power also does not change significantly across activation-frequency deciles, so it carries information beyond activation frequency.

Because neighbor crowding and the stability label both come from decoder cosine similarity, a crowded feature could clear $\theta$ by matching a nearby direction in the other seed rather than its true counterpart. To test this, we correlate each stable feature's activations on the evaluation tokens with those of its best cross-seed match. Matched pairs fire on the same tokens (median Pearson $r = 0.90$ at $k=64$ and $k=128$), while the second-best match, the nearest within-dictionary neighbor and a frequency-matched random feature do not (median $r \leq 0.03$). Agreement is highest in the most crowded decile, where the second-best match is closest ($0.95$ against $0.87$ in the most isolated at $k=64$), and requiring $r \geq 0.5$ in both partner seeds leaves the classifier essentially unchanged (AUROC $0.872$ at $k=64$).

\section{Discussion}
\subsection{Stability prediction}
These results suggest that whether a TopK latent will reappear under a different initialization is substantially predictable from six quantities measurable on the single run that produced it, at $\Delta = +0.259$ over the frequency baseline. The broader claim is that this predictability is a property of the training process rather than of a particular dictionary, which the near-lossless cross-seed transfer in \sectionref{sec:predictable} supports. Transfer across training budget is weaker. A classifier fitted at 50M tokens and applied at 1B scores $0.697$ with the source scaler ($0.630$ with the target's), about a third of the ceiling's gain over frequency alone. The statistics remain predictive within each checkpoint (AUROC $0.937$ at 50M), so the drop comes from fitting on a different population, with $15\%$ of labeled latents stable at 50M against $76\%$ at 1B.

\subsection{Cross-scale and cross-family transfer}
The cross-scale results show that stability prediction transfers across model sizes of the same family. Fitted on Pythia-70m, the classifier stays within $0.012$ of each Pythia target's ceiling with the target scaler, while beating each target's frequency baseline by more than $+0.26$. It also transfers to GPT-2 small at a larger cost ($0.051$), provided the statistics are standardized on the target, since a statistic's raw scale can differ across families. That step uses only the target's unlabeled statistics, so stability labels need to be paid for once rather than once per model.

\subsection{Crowding and feature stability}
\label{sec:crowded}
Our results on the neighbor crowding effect contradict the intuition that motivated this statistic. We expected an isolated direction to be one every seed is forced to find, and a crowded region to leave rotational freedom, yet we find that crowded features recur more than isolated ones across seeds.

One explanation of these results is that crowded regions represent genuine, important concepts that consistently demand model capacity across training runs, while isolated features may instead be idiosyncratic artifacts of initialization. This observation is broadly consistent with the Platonic Representation Hypothesis \citep{platonic_rep_hypothesis}, which proposes that neural networks converge toward a shared statistical model of reality. \citet{groger2026aristotelian} revisit this claim and find that, after calibration, the apparent convergence of global similarity measures largely disappears, while agreement in local neighborhood structure persists. They propose instead that representations converge to shared local neighborhood relationships. Our result has a similar flavor, since local density in the dictionary predicts which features recur. Their neighborhoods are defined over data points rather than dictionary atoms, however, so the parallel is an analogy rather than a direct test.

The second explanation follows from the reproducible-subspace account of \citet{gerasimov2026unstable}. If a feature lives in a space that consistently reproduces across seeds, then a different seeded SAE will contain \emph{some} direction in that space close enough to satisfy the similarity threshold $\theta = 0.7$, irrespective of it being the same feature. Our activation-overlap analysis (\sectionref{sec:confound}) argues against this, as we find that a feature's cross-seed match fires on the same tokens as the feature, most of all in crowded regions, while the next-closest direction does not, so the match recovers the feature itself rather than an arbitrary member of a shared subspace.

\section{Limitations}
\label{sec:limitations}

Experiments cover two model families, Pythia at three sizes (70m, 160m, 410m) and GPT-2 small, with Pythia-410m and GPT-2 trained to 500M tokens only, so the range of model capacity over which the effect holds is not established. Most results use three seeds, and a five-seed check at one setting ($k=64$, 500M tokens) does not weaken them. The classifier must be fitted once on labeled multi-seed data, carries to a second family only after re-standardization on the target, and is untested on other layers and SAE architectures. We keep the first sequence position, whose residual norm far exceeds the rest ($9\times$ the median for Pythia-70m, $34\times$ for GPT-2), and have not tested whether dropping it changes the results. It likely contributes to reconstruction contribution's larger scale in GPT-2. The Pythia-160m run was interrupted at 500M tokens and resumed from the start of the data stream, so its 1B budget is not 1B unique tokens, although all three seeds replayed the same data and cross-seed comparisons remain valid.

Another main limitation is that we feed the activation stream to every run in a fixed order, so each seed we run differs only in its initialization. While that matches prior work on SAE seed dependence \citep{gerasimov2026unstable}, it does not necessarily confirm whether stability survives a reordering of the data fed to it and would warrant further exploration. It also likely contributes to the high mean best-match cosine between our seeds ($0.79$ at 70m, $0.80$ at 160m, $0.75$ at 410m, $0.76$ at GPT-2 small), since seeds that share a data order are easier to match.

Finally, activation agreement establishes that matched features respond to the same inputs, but not that they play the same causal role in the model's computation; intervention-based analyses would be needed to establish the latter.

\section{Conclusion}
Cross-seed feature stability decreases as $k$ increases in TopK SAEs. The effect is not explained by dead features or within-dictionary redundancy, as, even when accounting for only those features which clear a dictionary-specific matching threshold, the stability of $k=256$ remains far lower than $k=64$ or $k=128$. Through several properties of a single run, we are able to substantially predict the stability of a given feature, with $\Delta = +0.259$ over frequency alone.

A classifier fitted on Pythia-70m predicts stability on Pythia-160m and Pythia-410m within $0.012$ of classifiers trained on their own labels, and on GPT-2 small, a different model family, within $0.051$ once its statistics are standardized on the target. The result also holds when stability is labeled from five seeds rather than three.
Neighbor crowding is the strongest single predictor of feature stability. While this analysis indicates that our classifier's performance shows success on these models, it remains an open question how well it generalizes more broadly.

\acks{This work was conducted as part of the Algoverse AI Research program.}

\bibliography{final_paper_bibliography}

\appendix

\section{Low stability at \texorpdfstring{$k=256$}{k=256}}
\label{apd:k256}
Relative to $k=64, 128$, we find that $k=256$ experiences near-zero stability. We explore and rule out a few simple explanations for that difference. One apparent explanation is that a significantly higher proportion of features in the latter case never fire; when looking at a training size of 1B tokens, $69.3\%$ of latents for $k=256$ are “dead” features, while the former see only $0.17\%$ and $1.4\%$, respectively. While, on the surface, this seems fully explanatory of the difference in cross-seed stability, in practice, it is not. When restricting cross-seed stability evaluation to only those latents which actually fire, feature stability for $k=256$ increases from $3.4\%$ to $\approx 16\%$, which is still significantly lower than that of $k=64$ ($\approx 64\%$) and of $k=128$ ($\approx 56\%$). As a result, the apparent issue of dead features is necessarily limited in its explanatory power of the result we observe.

Our fixed $\theta$, which we inherit directly for consistency with \citet{gerasimov2026unstable}, may be potentially miscalibrated, given that within-dictionary decoder similarity decreases as $k$ increases. We test this by recalibrating $\theta$ and building a null distribution for each $k$, computing maximum similarity against several random other features, then adopting the 95th percentile of that null as our unique threshold for the given value of $k$. This raises feature matching for $k=256$ from $1.0-4.4\%$ to $12.6-15\%$. Similarly to the first point, this is again significantly lower than $k=64, 128$, which have a range for matching of $65.7-83.5\%$, shown in \figureref{fig:similarity_distributions}. Separately, we also rule out near-duplicate features, which occur rarely across $k$ even when restricted to only looking at live features ($\approx 0\%$ for $k=64$, $k=128$, and $k=256$).

\begin{figure}[htbp]
\floatconts
  {fig:similarity_distributions}
  {\caption{Best-match cosine similarity vs null, by $k$ and training tokens, shows overlap at $k=256$ but clean separation at $k=64, 128$}}
  {\includegraphics[width=0.7\linewidth]
  {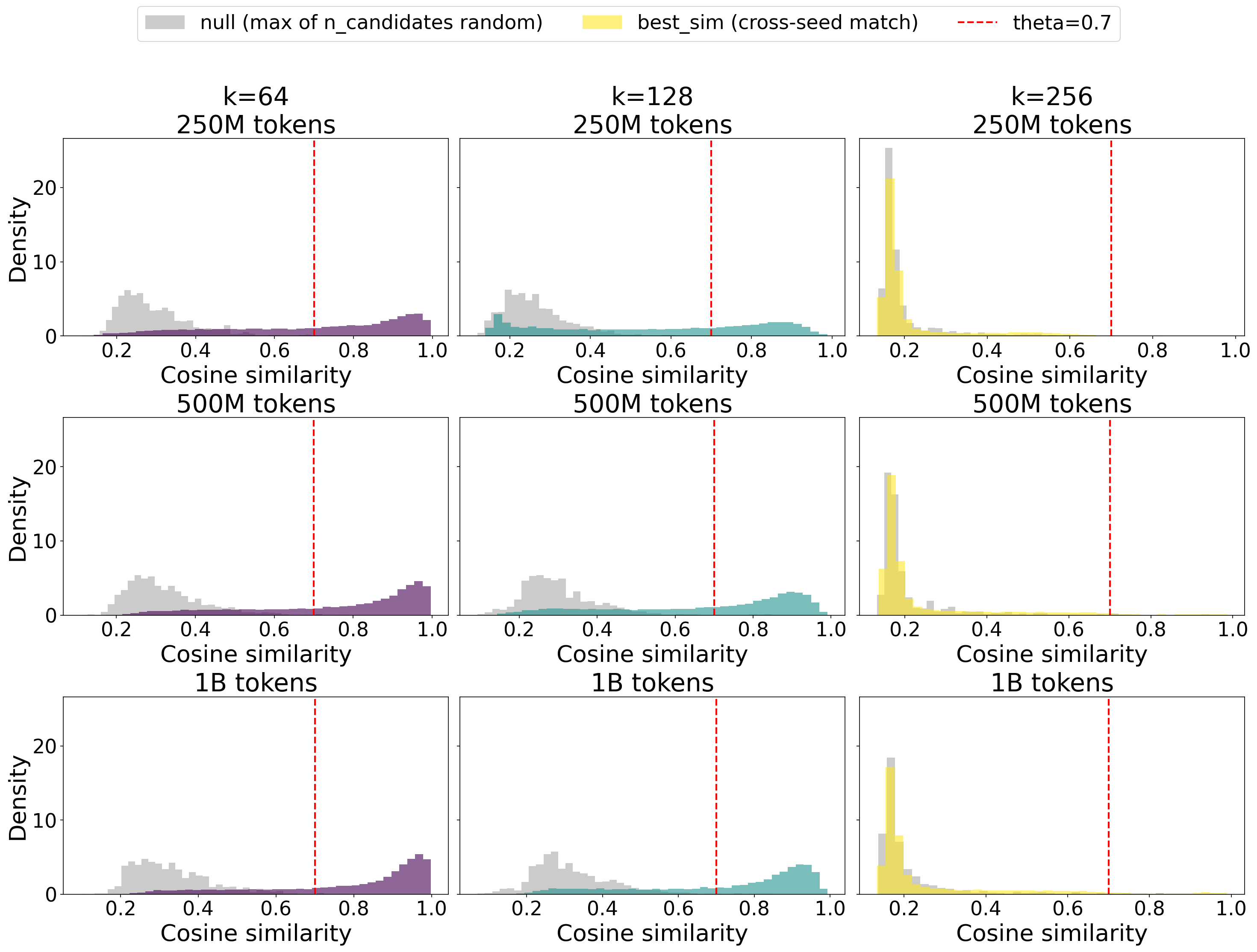}}
\end{figure}

\section{Classifier training}
\label{apd:training}
\paragraph{Balanced class weights.}
Stable and unstable features are not equally represented in our data, as the majority tend to be stable. This incentivizes the loss function to lean towards predicting ``stable,'' which results in minimal learning about the minority
class. Therefore, we weight each class inversely to its frequency,
\begin{equation}
  w_c = \frac{N}{2 N_c},
  \label{eq:classweight}
\end{equation}
where $N$ is the number of features and $N_c$ the number in class
$c \in \{\text{stable}, \text{unstable}\}$, so that the two classes contribute
equally to the total loss.

\paragraph{Stratification and five-fold cross-validation.}
We shuffle and separate the features that pass the floor threshold into five folds. Then we run five rounds of training and evaluation, rotating
which fold is held out and training on the remaining four, and report the mean
AUROC across rounds. Given the class imbalance, we also stratify the split, sampling from each class separately so that every fold
preserves the overall class ratio.

\section{Within-model replication}
\label{apd:by_model}
\tableref{tab:by_model} reports the classifier fitted and evaluated on each model's own labels.

\begin{table}[htbp]
  \centering
  \caption{Within-model replication at $k=64$, anchor seed 42, for the source model and the transfer targets. All rows use the single-anchor convention of \tableref{tab:by_k}.}
  \label{tab:by_model}
  \begin{tabular}{lccccccc}
    \toprule
    model & layer & $n$ & tokens & live & stable & AUROC & frequency only \\
    \midrule
    Pythia-70m  & 3 of 6   & $8192$     & 1B   & $97.3\%$ & $64.2\%$ & $0.868$ & $0.606$ \\
    Pythia-160m & 6 of 12  & $12{,}288$ & 1B   & $95.3\%$ & $67.2\%$ & $0.851$ & $0.573$ \\
    Pythia-410m & 12 of 24 & $16{,}384$ & 500M & $91.9\%$ & $56.2\%$ & $0.834$ & $0.562$ \\
    GPT-2 small & 6 of 12  & $12{,}288$ & 500M & $97.3\%$ & $57.2\%$ & $0.820$ & $0.492$ \\
    \bottomrule
  \end{tabular}
\end{table}

\section{Contribution of each statistic}
\label{apd:per_predictor}
\tableref{tab:per_predictor} reports, for each statistic, the AUROC of a classifier using that statistic alone, and the AUROC of the other five when it is removed.

\begin{table}[htbp]
  \centering
  \caption{Per-statistic contribution at $k=64$ and 1B tokens, averaged over three anchor seeds. \emph{alone}: AUROC using only that statistic; \emph{without}: AUROC of the other five; \emph{loss}: drop from the full six-statistic AUROC of $0.869$ when the statistic is removed.}
  \label{tab:per_predictor}
  \begin{tabular}{lccc}
    \toprule
    statistic & alone & without & loss \\
    \midrule
    Neighbor crowding           & $0.816$ & $0.837$ & $0.032$ \\
    Reconstruction contribution & $0.802$ & $0.867$ & $0.002$ \\
    Mean activation             & $0.780$ & $0.831$ & $0.037$ \\
    Encoder column norm         & $0.614$ & $0.860$ & $0.009$ \\
    Activation frequency        & $0.610$ & $0.869$ & $0.000$ \\
    Encoder bias                & $0.545$ & $0.865$ & $0.004$ \\
    \bottomrule
  \end{tabular}
\end{table}

\end{document}